\documentclass[11pt]{article}

\usepackage[preprint]{acl}
\usepackage[most]{tcolorbox}
\tcbuselibrary{breakable, skins} 
\usepackage{subfigure}
\usepackage{amsmath}
\usepackage{multirow}
\usepackage{amssymb} 
\usepackage{booktabs}
\usepackage{xcolor}
\usepackage{amssymb} 
\usepackage{tabularray}
\usepackage{longtable}
\usepackage{booktabs}
\usepackage{listings}
\usepackage{pifont}  
\usepackage{xcolor}  
\usepackage[dvipsnames]{xcolor}
\usepackage{supertabular}
\newcommand{\correct}{{\color{green}\ding{51}}} 
\newcommand{\wrong}{{\color{red}\ding{55}}}     
\NewDocumentCommand{\hanbin}
{ mO{} }{\textcolor{cyan}{\textsuperscript{\textit{hanbin}}\textsf{\textbf{\small[#1]}}}}
\NewDocumentCommand{\ganqu}
{ mO{} }{\textcolor{blue}{\textsuperscript{\textit{ganqu}}\textsf{\textbf{\small[#1]}}}}

\usepackage{times}
\usepackage{latexsym}

\usepackage[T1]{fontenc}

\usepackage[utf8]{inputenc}

\usepackage{microtype}

\usepackage{inconsolata}

\usepackage{graphicx}

\title{Teaching Large Reasoning Models Effective Reflection}

\author{
Hanbin Wang$^{1}$\thanks{Equal contribution.},
Jingwei Song$^{2}$\footnotemark[1],
Jinpeng Li$^{3}$\thanks{Corresponding author.},
Qi Zhu$^{3}$, \\\textbf{Fei Mi$^{3}$, Ganqu Cui$^{4}$, 
Yasheng Wang$^{}$, Lifeng Shang$^{3}$} \\
$^1$Peking University $^2$The University of Hong Kong \\ $^3$Huawei Technologies $^4$Shanghai AI Lab
}

\begin{document}
\maketitle

\begin{abstract}
Large Reasoning Models (LRMs) have recently shown impressive performance on complex reasoning tasks, often by engaging in self-reflective behaviors such as self-critique and backtracking. However, not all reflections are beneficial—many are superficial, offering little to no improvement over the original answer and incurring computation overhead. In this paper, we identify and address the problem of superficial reflection in LRMs. We first propose Self-Critique Fine-Tuning (SCFT), a training framework that enhances the model’s reflective reasoning ability using only self-generated critiques. SCFT prompts models to critique their own outputs, filters high-quality critiques through rejection sampling, and fine-tunes the model using a critique-based objective. Building on this strong foundation, we further introduce Reinforcement Learning with Effective Reflection Rewards (RLERR). RLERR leverages the high-quality reflections initialized by SCFT to construct reward signals, guiding the model to internalize the self-correction process via reinforcement learning. Experiments on two challenging benchmarks, AIME2024 and AIME2025, show that SCFT and RLERR significantly improve both reasoning accuracy and reflection quality, outperforming state-of-the-art baselines. 
All data and codes are available at \url{https://github.com/wanghanbinpanda/SCFT}.

\end{abstract}

\section{Introduction}
\begin{figure}[t] \centering
    \includegraphics[width=0.48\textwidth]{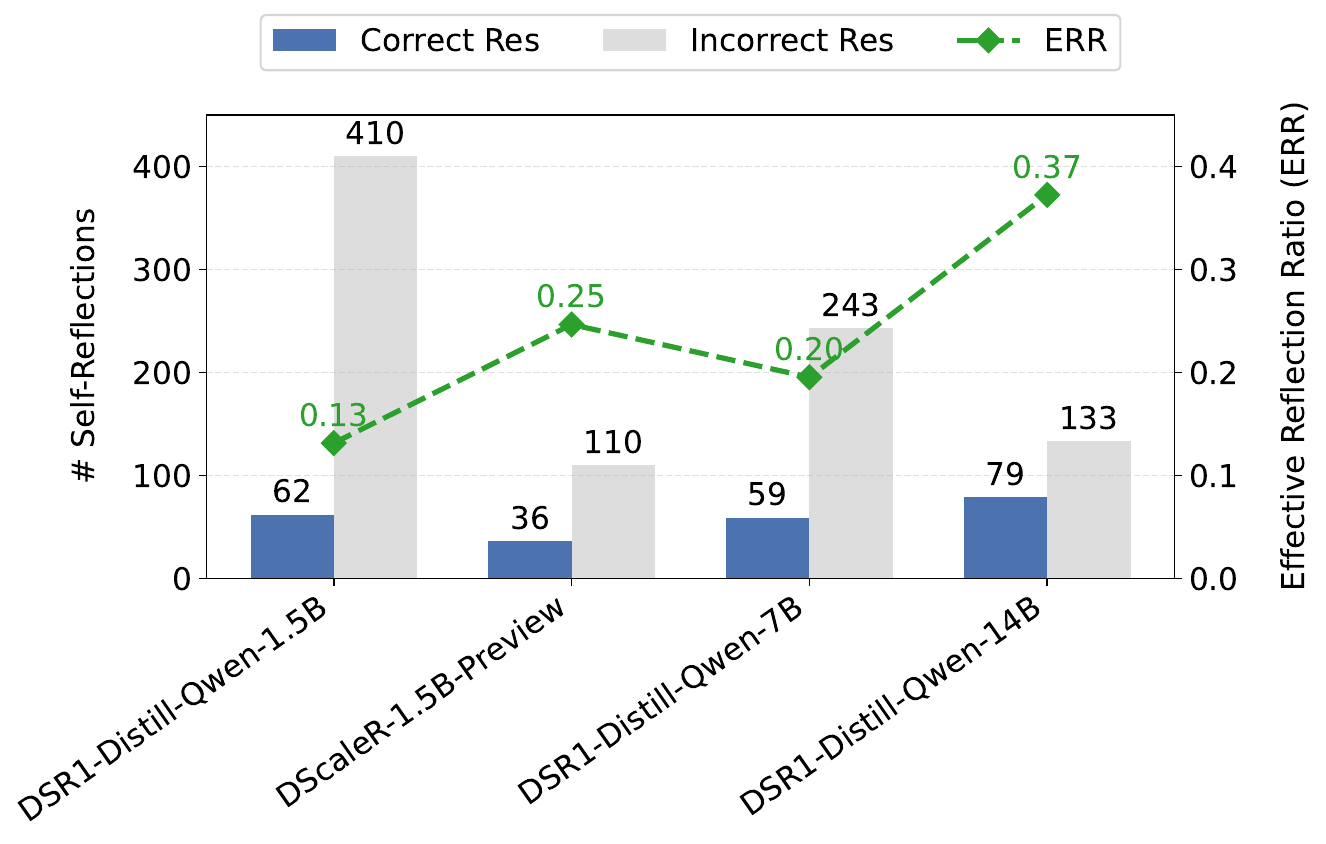}
    \caption{Total number of reflections in correct and incorrect responses on AIME2024. The blue and grey bars respectively show the total occurrences of reflection keywords in correct and incorrect responses. The Effective Reflection Ratio (ERR) is the proportion of reflection keywords in the correct responses, which is defined in Section \ref{model:def}. DS represents DeepSeek. Overall, the model is facing serious problems of ineffective reflection.} \label{fig:motivation}
\end{figure}


Recently, Large Reasoning Models (LRMs) such as DeepSeek-R1 \cite{deepseekai2025deepseekr1incentivizingreasoningcapability} and OpenAI o1 \cite{openai2024openaio1card} have demonstrated outstanding performance in solving complex reasoning tasks across domains such as mathematics, science, and programming. A key factor behind these advancements is the incorporation of reinforcement learning with verifiable rewards \cite{zeng2025simplerlzooinvestigatingtamingzero,cui2025processreinforcementimplicitrewards}, these models perform long Chain-of-Thought (CoT) \cite{wei2022chain} reasoning, engaging in detailed thinking before arriving at the final answer \cite{kimiteam2025kimik15scalingreinforcement,muennighoff2025s1simpletesttimescaling,ye2025limoreasoning}.


With the growing complexity of these reasoning traces, researchers have turned their attention to the thinking patterns that emerge within LRMs\cite{liu2025oatzero,wen2025thinkpatterns21ksystematicstudyimpact,gandhi2025cognitivebehaviorsenableselfimproving}. 
These studies reveal that LRMs exhibit a variety of reflective behaviors such as self-critique, verification, and backtracking—behaviors often associated with metacognition and expert reasoning. While such capabilities are desirable in principle, empirical evidence suggests that not all reflections are helpful (Figure \ref{fig:motivation}). Many reflections fail to lead to substantial answer improvements, and sometimes even introduce irrelevant or redundant information. We refer to this phenomenon as superficial reflection—a pattern of reevaluation that does not result in meaningful revision or better task performance. Superficial reflections not only undermine the effectiveness of reflection-based reasoning but also inflate inference cost and latency \cite{liu2025oatzero}.

To encourage more effective reflection, Critique Fine-Tuning (CFT) has been proposed as a promising training paradigm that fine-tunes models with critiques as supervision signals \cite{wang2025critiquefinetuninglearningcritique}. However, prior CFT approaches typically depend on high-quality critique annotations from stronger external models or human experts—resources that are expensive and difficult to scale\cite{lan2024traininglanguagemodelscritique,xie2025teachinglanguagemodelscritique,xi2024enhancingllmreasoningcritique}. In addition, previous attempts to use self-generated critiques for training have struggled to produce gains, due to the noisy and low-quality nature of these critiques\cite{wang2025critiquefinetuninglearningcritique}. Furthermore, these studies primarily focus on Short Chain-of-Thought (CoT) models, the critique finetuning for long CoT models has not been further explored yet.

In this paper, we first propose \textbf{Self-Critique Fine-Tuning (SCFT)}, a self-supervised framework that initializes effective reflective capabilities using only self-generated data. SCFT prompts the model to critique its own outputs, filters high-quality critiques via rejection sampling, and fine-tunes the model to internalize error correction. Building on this, we introduce \textbf{Reinforcement Learning with Effective Reflection Rewards (RLERR)} to further optimize the reasoning policy. Besides standard outcome-based RL, RLERR incorporates a hierarchical reward system based on effective reflection principles. Unlike sparse outcome rewards that only assess the final answer, RLERR provides dense feedback on the quality of the reflection process itself—rewarding truthfulness, constructiveness, and specificity. This guides the model to internalize the self-correction process and avoid superficial checks.

Our approach yields both performance improvements and behavioral gains. 
SCFT consistently enhances reasoning accuracy across various model scales. Compared with DeepSeek-R1-Distill-Qwen-7B, SCFT improves accuracy by 4.8\% and 6.0\% on AIME2024 and AIME2025, respectively. Even against the larger DeepSeek-R1-Distill-Qwen-14B, SCFT delivers consistent improvements of 0.9\% and 3.8\% on AIME2024 and AIME2025, respectively. Beyond accuracy, SCFT also enhances the model’s reflective behavior, increasing the Effective Reflection Ratio (ERR) by more than 10\% on average. Furthermore, RLERR pushes the boundary of effective reflection, enabling the DeepScaleR-1.5B-Preview model to achieve a Pass@1 of 44.2\% alongside a high Effective Reflection Ratio (ERR) of 0.48. Our analysis also reveals two critical insights: (1) SCFT serves as a superior initialization for reinforcement learning compared to standard self-distillation, unlocking higher performance ceilings; and (2) the synergy between reflection-quality rewards and outcome-based rewards is essential for guiding models to internalize genuine self-correction capabilities rather than superficial critiques.

\section{Related Work}

\subsection{Reasoning Behaviors of LRMs}
In recent years, Large Reasoning Models (LRMs) such as DeepSeek-R1 \cite{deepseekai2025deepseekr1incentivizingreasoningcapability} and OpenAI o1 \cite{openai2024openaio1card} have achieved superior performance in complex reasoning tasks. Through large-scale reinforcement learning, these reasoning models have learned longer Chain-of-Thought (CoT) thinking \cite{cui2025processreinforcementimplicitrewards,kimiteam2025kimik15scalingreinforcement,zeng2025simplerlzooinvestigatingtamingzero,deepscaler2025} and induce sophisticated reasoning behaviors such as reflection \cite{liu2025oatzero,gandhi2025cognitivebehaviorsenableselfimproving}. These reasoning behaviors of LRMs have attracted widespread attention from the research community. \citet{gandhi2025cognitivebehaviorsenableselfimproving} point out that LRMs primarily improve their performance during reasoning by employing reflective behaviors like verification and backtracking. ThinkPatterns \cite{wen2025thinkpatterns21ksystematicstudyimpact} investigates the impact of five thinking patterns, including self-critique, on the performance of LRMs. The self-critic pattern demonstrates good stability and scalability due to its two-stage generation and evaluation mechanism, offering a new perspective for enhancing the reasoning capabilities of LRMs. However, despite these reflective behaviors can enhance the reasoning ability of LRMs, \citet{liu2025oatzero} note that these models do not always engage in effective reflection, they often lack substantive correction or improvement of errors during the reflection process, which limits their ability to effectively enhance answer quality. Therefore, guiding LRMs to perform effective reflection has become an important direction for current research.


\subsection{Critique Learning}
Recent advances in critique learning have significantly enhanced the ability of LLMs to evaluate and improve model outputs. 
MultiCritique \cite{lan2024traininglanguagemodelscritique} leverages multi-agent feedback to generate high-quality critique data and employs reinforcement learning to further refine the model`s critique capabilities. 
AutoMathCritique \cite{xie2025teachinglanguagemodelscritique} focuses on automating the generation of step-level feedback for mathematical reasoning, significantly improving the model`s step-by-step critique performance on complex math problems through fine-tuning. 
CTRL \cite{xi2024enhancingllmreasoningcritique} teaches language models to critique via reinforcement learning, enabling them to dynamically adjust their critique strategies based on rewards. 
CFT \cite{wang2025critiquefinetuninglearningcritique} demonstrates that learning to critique is more effective than mere imitation, with models fine-tuned for critique tasks showing superior performance in generating high-quality and consistent outputs.
Nevertheless, these methods highly depend on stronger models or human experts for distillation or require a significant amount of time for reinforcement learning. Besides, previous work has not studied the effectiveness of critique fine-tuning on slow thinking models. Unlike them, SCFT focuses on the slow thinking reasoning models and constructs high-quality data through self-critique and rejection sampling to perform self-critique fine-tuning, enhancing both the model`s generation and critique capabilities simultaneously.

\subsection{Reinforcement Learning for Reasoning}
Reinforcement learning (RL) has become a cornerstone for aligning LLMs with complex reasoning tasks~\cite{deepseekai2025deepseekr1incentivizingreasoningcapability,openai2024openaio1card}. Early approaches primarily utilize Outcome Reward Models (ORMs), which provide sparse feedback based solely on the correctness of the answer~\cite{cobbe2021trainingverifierssolvemath}. While effective, ORMs often struggle with credit assignment in long reasoning chains. To address this, Process Reward Models (PRMs) were introduced to provide dense feedback by evaluating intermediate reasoning steps~\cite{lightman2023letsverifystepstep,wang2024mathshepherdverifyreinforcellms,cui2025processreinforcementimplicitrewards}. PRMs have shown significant improvements in mathematical reasoning by guiding models step-by-step. However, training PRMs typically requires expensive human annotations or supervision from super-sized models, limiting their scalability. Unlike standard PRMs that focus on step-level correctness, RLERR introduces a novel reward formulation centered on reflection quality. Instead of relying on external step-wise labels, we construct a hierarchical reward system based on effective reflection principles. This allows the model to internalize self-correction capabilities through reinforcement learning.



\section{Methodology}


\begin{figure*}[t] \centering
    \includegraphics[width=1\textwidth]{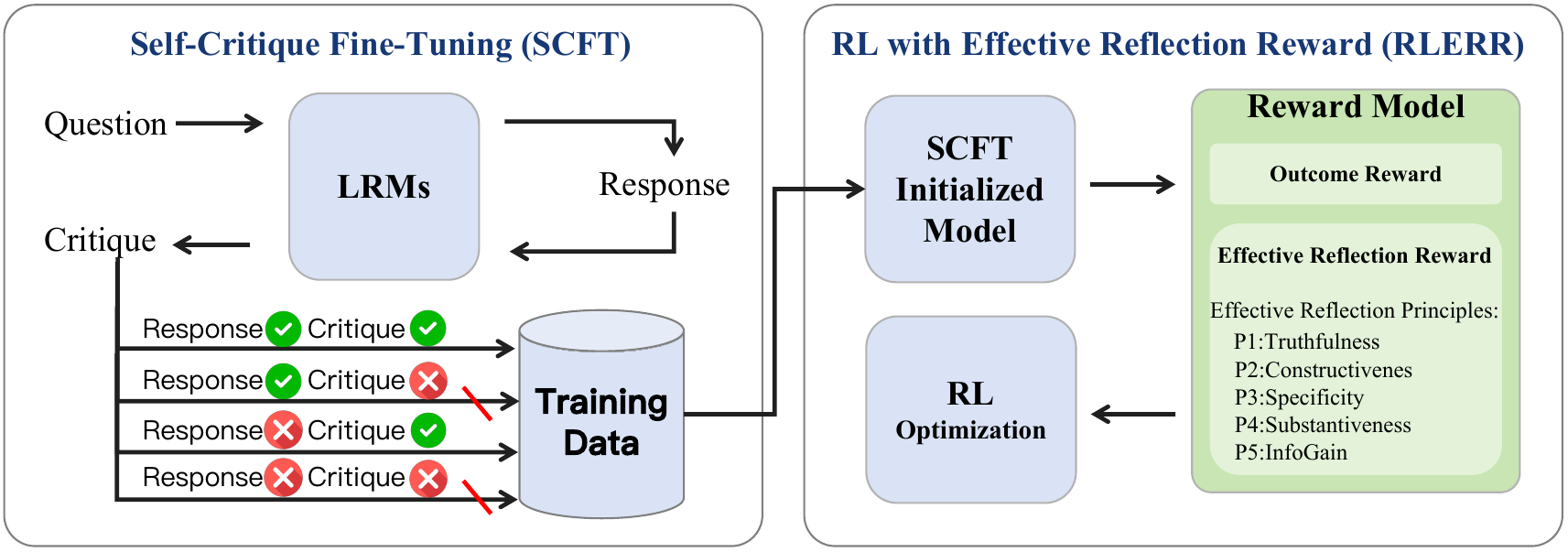}
    \caption{The overall pipeline of our proposed SCFT and RLERR method.} \label{fig:scft_rl}
\end{figure*}
In this section, we present a systematic framework for enhancing the reflective capabilities of Large Reasoning Models (LRMs). Our methodology consists of three main components. First, in Sec.~\ref{model:def}, we define the concept of Effective Reflection and its quantitative metric. 
Then, as illustrated in Figure~\ref{fig:scft_rl}, we first propose Self-Critique Fine-Tuning (SCFT) (Sec.~\ref{model:scft}) and then introduce Reinforcement Learning with Effective Reflection Rewards (RLERR)  (Sec.~\ref{sec:rl_rewards}). SCFT is a supervised learning approach that leverages high-quality self-generated critiques to equip the model with fundamental reflective skills. RLERR incentivizes the model to generate deep reflections that effectively correct errors, optimizing the model's policy using the defined reward signals.

\subsection{Effective Reflection}\label{model:def}
Effective reflection is the process by which a large reasoning model identifies and corrects errors in its reasoning or confirms that the reasoning is correct, leading to improved solutions.
Let $x$ represent a question and $y$ represent a model's response. We consider a reflection $r \subset y$ to be effective if it contributes to arriving at a correct solution. 

Previous work\cite{liu2025oatzero} uses the number of reflection keywords to generally reveal that the reflection of LRMs is not always effective. These reflections might introduce errors into originally correct answers or fail to produce a valid answer after several reflections. In order to see more clearly the proportion of effective reflection in the model reasoning process and measure reflection effectiveness, we introduce the Effective Reflection Ratio (ERR), defined as the proportion of reflections that contribute to correct solutions:

\begin{equation}
\text{ERR} = \frac{\sum_{i=1}^n \text{Accuracy}(y_i) \cdot R(y_i)}{\sum_{i=1}^n R(y_i)}
\label{eq:err}
\end{equation}
where $n$ is the total number of responses, $y_i$ denotes the $i$-th response, $R(y_i)$ counts the number of reflections in response $y_i$, and $\text{Accuracy}(y_i) \in \{0,1\}$ is a binary indicator of whether $y_i$ is fully correct. 


\subsection{Self-Critique Fine-Tuning (SCFT)}\label{model:scft}

Self-Critique Fine-Tuning (SCFT) aims to teach large reasoning models to effectively reflect. In this subsection, we present the construction of training data using self-critique (Sec.~\ref{model:traing_data}), training objective, and inference mode (Sec.~\ref{model:traing_objective}).

\subsubsection{Construction of Training Data}\label{model:traing_data}

Given a question $q$, the model $\mathcal{M}$ first generates a response $y$, followed by a critique $c$ that evaluates the correctness of $y$:

\begin{equation}
y=\mathcal{M}\left(q\right), c=\mathcal{M}\left(t{\|q\|}y\right)
\end{equation}
where $t$ denotes a predefined critique prompt template, and $\|$ represents sequence concatenation. We then employ ground truth based rejection sampling to filter high-quality data as in previous work \cite{openr1,openthoughts}. A generated triplet $(q, y, c)$ is included in the training set if and only if it satisfies the following correctness filter:

\begin{equation}
\mathcal{F}(q, y, c) =
\begin{cases}
1, & \text{if } \text{Acc}(y) = 1 \text{ and } \text{Acc}(c) = 1, \\
1, & \text{if } \text{Acc}(y) = 0 \text{ and } \text{Acc}(c) = 1, \\
0, & \text{otherwise},
\end{cases}
\end{equation}
where $\text{Acc}(\cdot)$ evaluates correctness against the ground truth. This filtering ensures that only samples with valid self-reflection behavior are retained. The full training dataset is then defined as:

\begin{equation}
\mathcal{D} = \left\{ (q_i, y_i, c_i) \mid \mathcal{F}(q_i, y_i, c_i) = 1 \right\}.
\end{equation}

Importantly, all responses and critiques are self-generated by the same model $\mathcal{M}$, without reliance on external supervision or stronger teacher models. Both outputs adopt a long-form Chain-of-Thought (CoT) format, consisting of an explicit reasoning trace enclosed by "<think>" and "</think>" tags, followed by a concise answer summary. Examples of accepted training instances are shown in Appendix \ref{app:example}. 


\subsubsection{Training Objective}\label{model:traing_objective}
During the fine-tuning phase, we train the model on the constructed dataset $D$, where the input is ``Please critique whether the following solution to the question is correct. Question: \{$q$\} Solution: \{$y$\}'', and the expected output is the critique $c$. The training objective is to minimize the negative log-likelihood:

\begin{equation}
\mathcal{L} = -\sum_{(q,y,c) \in D} \log P(c|q,y;\theta)
\end{equation}
where $\theta$ represents the model parameters, $q$ is the question, $y$ is the response, and $c$ is the critique.

At inference time, we can regard the trained model as both a generative model (producing answers) and a critique model (evaluating and improving answers) simultaneously by simply modifying the prompt. For the generative mode, we input the question $q$ and obtain the response $y$. For the critique mode, we input the question $q$ along with the generated response $y$ to produce the critique $c$. In particular, we did not use additional question-answer pairs to train the model.

After training, the model can act as a generative model or a critique model by modifying the prompt. While SCFT equips the model with preliminary reflective capabilities, to further enhance the accuracy and robustness of reflection, we utilize the SCFT-tuned model as the initialization for the subsequent reinforcement learning stage.

\subsection{Reinforcement Learning with Effective Reflection Rewards (RLERR)}
\label{sec:rl_rewards}

While SCFT provides a strong initialization for reflective reasoning, supervised fine-tuning alone is limited by the quality of static data. To further align the model with Effective Reflection, we introduce RLERR, which is reinforcement learning with a hierarchical reward system based on five core principles of reflection. 

\subsubsection{Hierarchical Reflection Principles}
We define five principles ($P_1$ to $P_5$) for evaluating reflection quality, ranked from fundamental validity to advanced reasoning optimization. Lower-level principles act as prerequisites for higher-level rewards.
\begin{itemize}
    \item \textbf{$P_1$: Truthfulness (Critical Necessity).} The reflection must objectively judge the correctness of the previous step. It should not hallucinate errors in correct steps nor blindly validate incorrect ones.
    \item \textbf{$P_2$: Constructiveness (Problem Solving).} The reflection must not only identify an error but also propose a valid correction or a specific next step. Mere complaints (e.g., ``This is wrong'') without guidance are penalized.
    \item \textbf{$P_3$: Specificity (Precision).} The critique must pinpoint the exact location or logic of the error (e.g., ``calculation error in step 2'' vs. ``the answer looks wrong'').
    \item \textbf{$P_4$: Substantiveness (Depth).} The reflection should involve rigorous verification (e.g., re-deriving a formula or checking constraints) rather than superficial checks.
    \item \textbf{$P_5$: InfoGain (Efficiency).} The reflection should break repetitive loops or open new reasoning paths that significantly reduce uncertainty, avoiding redundant restatements.
\end{itemize}

\subsubsection{Training Setup}
We employ GRPO as our RL algorithm and use the SCFT-initialized model as the policy model $\pi_\theta$. 
To obtain dense and scalable reward signals, we utilize a strong LLM (e.g., GPT-4o) as the \textbf{Reward Model}. The Reward Model evaluates the generated trajectory based on the five principles defined above and assigns a holistic scalar score $R \in [0, 10]$. 
The scoring process is guided by a comprehensive prompt (detailed in Appendix \ref{app:evaluation_prompt}) that instructs the judge to heavily penalize hallucinations (violating $P_1$) while rewarding constructive and specific insights ($P_2$-$P_5$). This approach allows the optimization process to capture the nuance of effective reflection without rigid rule-based filtering. We employ GRPO to maximize the expected reward:
\begin{equation}
\max_\theta \mathbb{E}_{(q) \sim D, y \sim \pi_\theta} [R(q, y)]
\end{equation}
where $R(q, y)$ is the score provided by the LLM judge.

\section{Experimental Methodology}
In this section, we describe the datasets, evaluation metrics, baselines, and implementation details.

\textbf{Dataset. }For the SCFT training set, we use the mathematical problems provided in DeepScaleR\cite{deepscaler2025} to construct the data through self-critique. For the RL training set, we usethe  DAPO-Math-17K dataset \citep{yu2025dapoopensourcellmreinforcement}, which is a curated collection of approximately 17,000 competition-level math problems. For testing, we evaluate the effectiveness of Self-Critique Fine-Tuning on AIME2024 \cite{numina_math_datasets}, AIME2025 \cite{ye2025limoreasoning}, MATH-500 \cite{hendrycksmath2021}, and GPQA datasets \cite{rein2023gpqagraduatelevelgoogleproofqa}. 


\textbf{Evaluation Metrics. }We follow previous work~\cite{chen2021evaluatinglargelanguagemodels,li2024mmcode,wang2024intervenor,yang2024enhancing,luo2023wizardcoder} and we use \texttt{Pass@k} ~\cite{chen2021evaluatinglargelanguagemodels} to evaluate the effectiveness of different models. \texttt{Pass@k} represents the probability that at least one correct solution appears among the top $k$ generated solutions for each problem:
\begin{equation}
    \text{Pass@}k:=\underset{\text{Problems}}{\operatorname*{\mathbb{E}}}\left[1-\frac{\binom{n-c}k}{\binom nk}\right]
\end{equation}
where $n$ denotes the total number of generated solutions, $c$ is the number of correct solutions, and $k$ is the number of top-ranked solutions being evaluated. In this work, we set $k=1$.
The \texttt{Pass@1}  accuracy is averaged over 16 samples per problem.
Besides, we use \texttt{ERR} to evaluate the proportion of effective reflection in the response.

\begin{table*}[t]
\centering
\resizebox{\linewidth}{!}{
\begin{tabular}{lcccccccccc}
\hline
\multirow{2}{*}{\textbf{Model}} & \multicolumn{2}{c}{\textbf{AIME2024}}                & \multicolumn{2}{c}{\textbf{AIME2025}}                & \multicolumn{2}{c}{\textbf{MATH-500}}                & \multicolumn{2}{c}{\textbf{GPQA Diamond}}            & \multirow{2}{*}{\textbf{Avg. Pass@1}} & \multirow{2}{*}{\textbf{Avg. ERR}} \\ \cline{2-9}
                                & \multicolumn{1}{l}{Pass@1} & \multicolumn{1}{l}{ERR} & \multicolumn{1}{l}{Pass@1} & \multicolumn{1}{l}{ERR} & \multicolumn{1}{l}{Pass@1} & \multicolumn{1}{l}{ERR} & \multicolumn{1}{l}{Pass@1} & \multicolumn{1}{l}{ERR} &                                       &                                    \\ \hline
Qwen-2.5-7B-SimpleRL-Zoo                 & 15.0                       & 0.06           & 5.2                       & 0.03                    & 69.1              & 0.26           & 31.3                       & 0.14                    & 30.1                         & 0.12                      \\ 
Light-R1-7B-DS                 & 57.7                       & 0.32           & 46.4                       & 0.23                    & 91.1              & 0.75           & 47.2                       & 0.44                    & 60.6                         & 0.44                      \\ 
AReaL-boba-RL-7B                 & \textbf{62.7}                       & \textbf{0.39}           & \textbf{49.4}                       & \textbf{0.24}                    & 93.8              & 0.71           & 48.0                       & 0.45                    & 63.5                         & \textbf{0.45}                      \\ 
LIMO-32B                 & 55.2                       & 0.24           & 44.8                       & 0.19                    & \textbf{94.8}              & \textbf{0.77}           & \textbf{63.1}                       & \textbf{0.50}                    & \textbf{64.5}                         & 0.43                      \\ 
s1-32B                 & 50.2                       & 0.22           & 26.0                       & 0.15                    & 92.6              & 0.67           & 57.0                       & 0.47                    & 56.5                         & 0.38                      \\ 
\hline

DeepSeek-R1-Distill-Qwen-1.5B   & 29.0                       & 0.13                    & \textbf{23.3}              & \textbf{0.14}           & 82.6                       & 0.46                    & \textbf{33.3}              & \textbf{0.34}           & 42.0                                  & 0.27                               \\
+ Self-Distill                  & \textbf{32.3}              & \textbf{0.15}           & 21.9                       & 0.06                    & 81.4                       & 0.50                    & 31.1                       & 0.30                    & 41.7                                  & 0.25                               \\
+ SCFT                 & 31.9                       & \textbf{0.15}           & 22.9                       & 0.13                    & \textbf{82.6}              & \textbf{0.53}           & 32.7                       & 0.33                    & \textbf{42.5}                         & \textbf{0.29}                      \\ \hline
DeepScaleR-1.5B-Preview         & 40.2                       & 0.32                    & 28.5                       & 0.17                    & 87.7              & \textbf{0.77}           & 32.3              & 0.41                    & 47.2                                  & 0.42                               \\
+ Self-Distill                  & 37.3                       & 0.43           & 29.8              & 0.19           & 87.2                       & 0.70                    & 30.2                       & 0.41           & 46.1                                  & 0.43                             \\
+ SCFT                 & 41.3              & 0.40                    & 28.3                       & 0.19           & 87.3                       & 0.71                    & 32.3             & 0.43           & 47.3                         & 0.43   \\
\hspace{0.3cm}+ RLERR                 &  \textbf{44.2}             &  \textbf{0.48}                   &   \textbf{31.0}                     &  \textbf{0.33}          &   \textbf{89.5}                     &   0.76                  &   \textbf{37.9}            &  \textbf{0.50}          & \textbf{50.7}                         & \textbf{0.52}                      \\ \hline
DeepSeek-R1-Distill-Qwen-7B     & 55.6                       & 0.26                    & 39.4                       & 0.14                    & 91.9                       & 0.65                    & 47.1                       & 0.46           & 58.5                                  & 0.38                               \\
+ Self-Distill                  & 57.9                       & 0.25                    & 40.2                       & 0.19                    & 89.8                       & 0.63                    & 48.5                       & 0.42                    & 59.1                                  & 0.37                               \\
+ SCFT                 & 60.2              & 0.38           & 45.4              & 0.21           & \textbf{92.4}              & 0.72           & 48.7              & 0.40                    & 61.7                         & 0.43                   \\
\hspace{0.3cm}+ RLERR                 &  \textbf{63.1}             &   \textbf{0.50}                  &   \textbf{48.1}                    &  \textbf{0.33}          &   92.2                     &   \textbf{0.80}                  &   \textbf{57.0}            &  \textbf{0.51}          &    \textbf{65.1}                      & \textbf{0.54}    \\ \hline
DeepSeek-R1-Distill-Qwen-14B    & 70.4                       & 0.52           & 50.0                       & 0.27                    & 92.4                       & 0.68                    & 59.5                       & \textbf{0.58}           & 68.1                                  & 0.51                               \\
+ Self-Distill                  & 70.4                       & 0.52           & 52.1                       & 0.35                    & 92.4                       & 0.73                    & 59.5                       & 0.53                    & 68.6                                  & 0.53                               \\
+ SCFT                 & \textbf{71.3}              & \textbf{0.53}                    & \textbf{53.8}              & \textbf{0.47}           & \textbf{93.4}              & \textbf{0.77}           & \textbf{60.8}              & 0.53                    & \textbf{69.8}                         & \textbf{0.57}                      \\ \hline
\end{tabular}
}
\caption{Overall Performance of Self-Critique Fine-Tuning. The \texttt{Pass@1} is averaged over 16 samples per problem. Due to the limitations of computing resources, we only conduct RLERR on DeepScaleR-1.5B-Preview and DeepSeek-R1-Distill-Qwen-7B.}
\label{tab:overall}
\end{table*}

\textbf{Baselines. }In our experiments, we perform self-critique fine-tuning on DeepSeek-R1-Distill-Qwen-1.5B/7B/14B\cite{deepseekai2025deepseekr1incentivizingreasoningcapability} and DeepScaleR-1.5B-Preview \cite{deepscaler2025}, and compare it with the self-distillation method. The data for self-distillation is generated by the model itself and filtered using ground truth to obtain high-quality data. The amount of self-distill data is the same as that of SCFT data. Furthermore, we also make comparisons with some advanced models, such as Qwen-2.5-7B-SimpleRL-Zoo\cite{zeng2025simplerlzooinvestigatingtamingzero}, Light-R1-7B-DS\cite{wen2025lightr1curriculumsftdpo}, AReaL-boba-RL-7B\cite{areal2025}, LIMO-32B\cite{ye2025limoreasoning}, and s1-32B\cite{muennighoff2025s1simpletesttimescaling}.

\textbf{Implementation Details. }During SCFT stage, all models are trained using the Llama-Factory framework \cite{zheng2024llamafactory}. We set the learning rate to 1e-5, the number of training epochs to 10, and the batch size to 64. 
During RL, we use GRPO as the RL algorithm. For hyperparameters, we set the batch size and mini-batch size to $64$, and for each problem, we rollout $8$ responses. The maximum lengths for prompts and responses are $1,024$ and $16,384$ tokens, respectively. The learning rate is set to $1e-6$, and we adopt the AdamW optimizer for the policy model. During inference, we set the temperature to 0.6 and the maximum generation length to 32,768 tokens.

\section{Evaluation Results}
In this section, we evaluate the overall performance of Self-Critique Fine-Tuning (SCFT) and RLERR. Then we conduct ablation studies and also explore the influence of using different critique models on the performance of SCFT. Finally, we demonstrate that SCFT serves as a superior initialization for RL and RLERR (combining outcome and effective reflection rewards) significantly outperforms using outcome or reflection rewards in isolation.

\subsection{Overall Performance}
\begin{figure*}[t] \centering
    \includegraphics[width=0.96\textwidth]{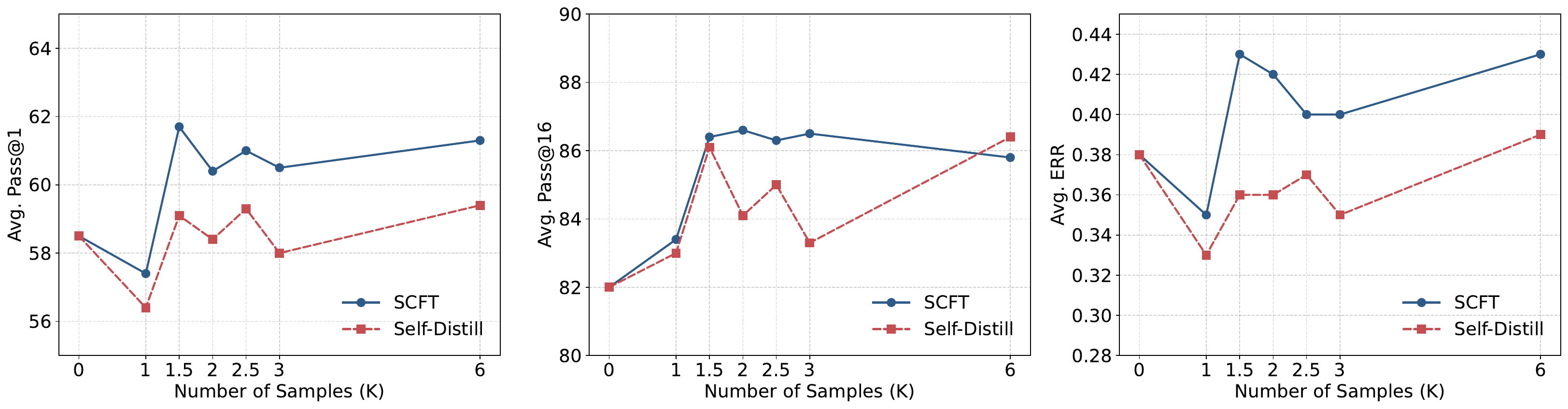}
    \caption{The impact of the amount of SCFT and Self-Distill training data on the model performance. The base model is DeepSeek-R1-Distill-Qwen-1.5B. The \texttt{Pass@1} represents the average performance across all test sets.} \label{fig:number_of_samples}
\end{figure*}
The overall performance of SCFT and RLERR is shown in Table \ref{tab:overall}. SCFT consistently improves reasoning performance across models of varying scales. The 7B and 14B variants of DeepSeek-R1-Distill-Qwen achieve the most significant gains: SCFT boosts average \texttt{Pass@1} by 2.6\% (58.5→61.7) and 1.7\% (68.1→69.8), respectively, outperforming self-distillation baselines. Additionally, the effective reflection ratios of the 7B and 14B models increase by 5\% and 6\%, respectively, indicating that the models achieve more accurate reflection through SCFT. Notably, on the more challenging datasets AIME2024 and AIME2025, the 7B model's \texttt{Pass@1} improves by 4.6\% and 6\%, respectively, demonstrating that SCFT can further enhance the model's reasoning capabilities. Moreover, on the 1.5B model, SCFT achieves marginal gains on \texttt{Pass@1}. For DeepScaleR-1.5B-Preview, SCFT brings a 4.3\% improvement in \texttt{Pass@16} (Figure \ref{fig:pass16}), suggesting that SCFT can raise the upper limit of the model's performance. Additionally, the 7B models after SCFT achieve comparable results with the advanced 7B models SimpleRL-Zoo, Light-R1-7B-DS, and AReaL-boba-RL-7B, which have undergone extensive reinforcement learning. Moreover, it outperforms the larger-scale s1-32B model and achieves comparable results with LIMO-32B.

RLERR yields further improvements upon SCFT, significantly enhancing both reasoning accuracy and reflection quality. As shown in Table \ref{tab:overall}, applying RLERR to DeepScaleR-1.5B-Preview results in an average \texttt{Pass@1} of 50.7\% and an average ERR of 0.52, surpassing the SCFT stage by 3.4\% and 0.09, respectively. Similarly, for DeepSeek-R1-Distill-Qwen-7B, RLERR pushes the average \texttt{Pass@1} to 65.1\%, demonstrating a substantial gain over the SCFT baseline (61.7\%). Notably, the 7B model trained with RLERR achieves state-of-the-art performance among models of similar size, outperforming AReaL-boba-RL-7B (63.5\%) and Light-R1-7B-DS (60.6\%). Furthermore, it even surpasses the 32B-scale model LIMO-32B (64.5\%) in average accuracy.




\subsection{Ablation Studies}

\begin{table}[t]
\centering
\resizebox{\linewidth}{!}{
\begin{tabular}{lccccc}
\hline
\textbf{Model}                     & \textbf{\# c-\textgreater c} & \textbf{Pass@1}                      & \textbf{Pass@16}                     & \textbf{ERR}  \\ \hline
DS-R1-Dis-Qwen-7B                                   & -                             & 58.5                                 & 82.0                                 & 0.38          \\ \hline
                                                             & 0.0                           & 60.8                                 & 84.9                                 & 0.38          \\
                                       & {\color[HTML]{000000} 0.5k}   & {\color[HTML]{000000} \textbf{61.7}} & {\color[HTML]{000000} \textbf{86.4}} & \textbf{0.43} \\
                                                              & 1k                            & 60.7                                 & 85.0                                 & 0.38          \\
                                                              & 1.5k                          & 60.3                                 & 85.0                                 & 0.41          \\
                                                              & 2k                            & 60.8                                 & 86.3                                 & 0.40          \\
                                                              & 2.5k                          & 60.2                                 & 84.7                                 & 0.37          \\
\multirow{-7}{*}{+ SCFT (\# i->c=1k)}                             & 3k                            & 60.2                                 & 85.8                                 & 0.39          \\ 

\hline

\end{tabular}
}
\caption{The Trade-off Between Error Correction (i→c) and Correctness Reinforcement (c→c) in Self-Critique Fine-Tuning. \# i→c denotes the number of critiques correcting incorrect responses, while \# c→c represents the number of critiques validating correct responses. The experimental results represent the average performance across all test sets.}
\label{tab:ratio}
\end{table}

To further investigate the effect of the proportion and amount of data used for self-Critique fine-tuning (SCFT) on the performance of fine-tuning, we perform the ablation Studies.
The data used for SCFT can be categorized into two types: \texttt{i→c} and \texttt{c→c}. Specifically, \texttt{i→c} represents critiques that identify the mistakes in incorrect responses and give the correct answer, while \texttt{c→c} represents critiques that confirm correct responses. As shown in Table \ref{tab:ratio}, we investigate the impact of the ratio of these two types of data on model performance. By fixing the amount of \texttt{i→c} data at 1k and varying the amount of \texttt{c→c} data, we observe that the optimal performance is achieved when the ratio of \texttt{i→c} to \texttt{c→c} is 2:1.

Following this optimal ratio, we further explore the influence of the total amount of critique data used for SCFT on model performance. As shown in Figure \ref{fig:number_of_samples}, when scaling the training data from 1K to 6K samples, SCFT provides a greater performance improvement than self-distillation. However, as the amount of data increases, the performance gains of SCFT and self-distillation gradually level off. This indicates that excessively increasing the data for SCFT and self-distillation does not lead to a continuous improvement in model performance.

\begin{table}[t]
\centering
\small
\resizebox{\linewidth}{!}{
\begin{tabular}{lccc}
\hline
Model                       & Pass@1        & Pass@16       & ERR           \\ \hline
DPSK-R1-Dis-Qwen-7B & 55.6          & 83.3          & 0.26          \\
+ WebInstruct-CFT           & 44.8          & 80.0          & 0.21          \\
+ Self-Distill FT             & 57.9          & 86.7          & 0.25          \\
+ R1-Distill FT              & 59.2          & 86.7          & 0.41          \\
+ Self-Critique FT             & \textbf{60.2} & 86.7          & 0.38          \\
+ R1-Critique FT               & \textbf{60.2} & \textbf{90.0} & \textbf{0.49} \\ \hline
\end{tabular}
}
\caption{The Performance Comparison of Critique Fine-Tuning Using Different Critique Models on AIME 2024. FT represents Finetuning.}
\label{tab:teacher_model}
\end{table}
\begin{table*}[t]
\centering
\footnotesize
\resizebox{0.8\linewidth}{!}{
\begin{tabular}{lccccc}
\hline
\textbf{Critique Model} & \textbf{Accuracy@t1} & \textbf{Accuracy@t2} & \textbf{$\Delta$(t1, t2)} & \textbf{$\Delta_{i \to c}$(t1, t2)} & \textbf{$\Delta_{c \to i}$(t1, t2)} \\ \hline
DPSK-R1-Dis-Qwen-7B     & 23.3    & 20.0          & -3.3                & 3.3               & 6.7               \\
+ Self-Critique FT        & 23.3    & 33.3          & 10.0                & 13.3              & \textbf{3.3}   \\
\hspace{0.3cm}+ RLERR        & 23.3    & \textbf{50.0} & \textbf{26.7}       & \textbf{30.0}     & \textbf{3.3}   \\

+ R1-Critique FT        & 23.3    & 40.0 & 16.7       & 20.0     & \textbf{3.3}      \\ \hline
\end{tabular}
}
\caption{The Critique Performance of Critique Fine-Tuning on AIME 2024. The initial response generation model is Qwen2.5-Math-7B-Instruct. Accuracy@t1 represents the accuracy of the initial generation. Accuracy@t2 represents the accuracy of the critique model after one round of correction. $\Delta_{i\to c}$(t1, t2) indicates the fraction of problems that are incorrect in the first generation but become correct after critique. $\Delta_{c\to i}$(t1, t2) indicates the fraction of problems that are correct in the first generation but become incorrect after critique. FT represents Finetuning.}
\label{tab:self-correction}
\end{table*}
\subsection{Performance Comparison of CFT Using Different Critique Models}
We investigate the impact of using a stronger model as a critique model to construct data for Critique Fint-Tuning on model performance.
The results are shown in Table \ref{tab:teacher_model}. WebInstruct-CFT is a critique-based instruction dataset derived from WebInstruct. It includes critiques of responses, and the responses and critiques are all in the form of Short Chain-of-Thought (short CoT). Self/R1-distill and Self/R1-Critique are the data for SFT and SCFT constructed by self and DeepSeek-R1, respectively.  From the results, we observe that fine-tuning with data constructed by a stronger model can achieve better performance. Specifically, when using R1-Critique finetuning,  both the \texttt{Pass@1} and \texttt{Pass@16} metrics achieve the highest scores. Furthermore, for the Long CoT model, if short CoT data (WebInstruct-CFT) is used for training, the performance will be reduced.

We further explore the critique ability of the model, and the results are shown in Table \ref{tab:self-correction}. We use Qwen2.5-Math-7B-Instruct\cite{yang2024qwen25mathtechnicalreportmathematical} as the generative model to generate the initial response and use different models for critique. From the results, we can see that DeepSeek-R1-Distill-Qwen-7B has a poor ability to correct incorrect responses and cannot distinguish the correctness of responses, often modifying correct responses to incorrect ones. In contrast, the model`s critique ability is significantly enhanced after SCFT and RLERR, and the model has stronger discrimination capabilities, avoiding the modification of correct responses to incorrect ones. Moreover, by using a stronger model, DeepSeek-R1, to construct SCFT data, the model`s critique ability is further improved compared to self-critique.

\begin{figure}[t]
    \centering
    \subfigure[Impact of Initialization Strategies.] { \label{subfig:rl1} 
    \includegraphics[width=0.48\linewidth]{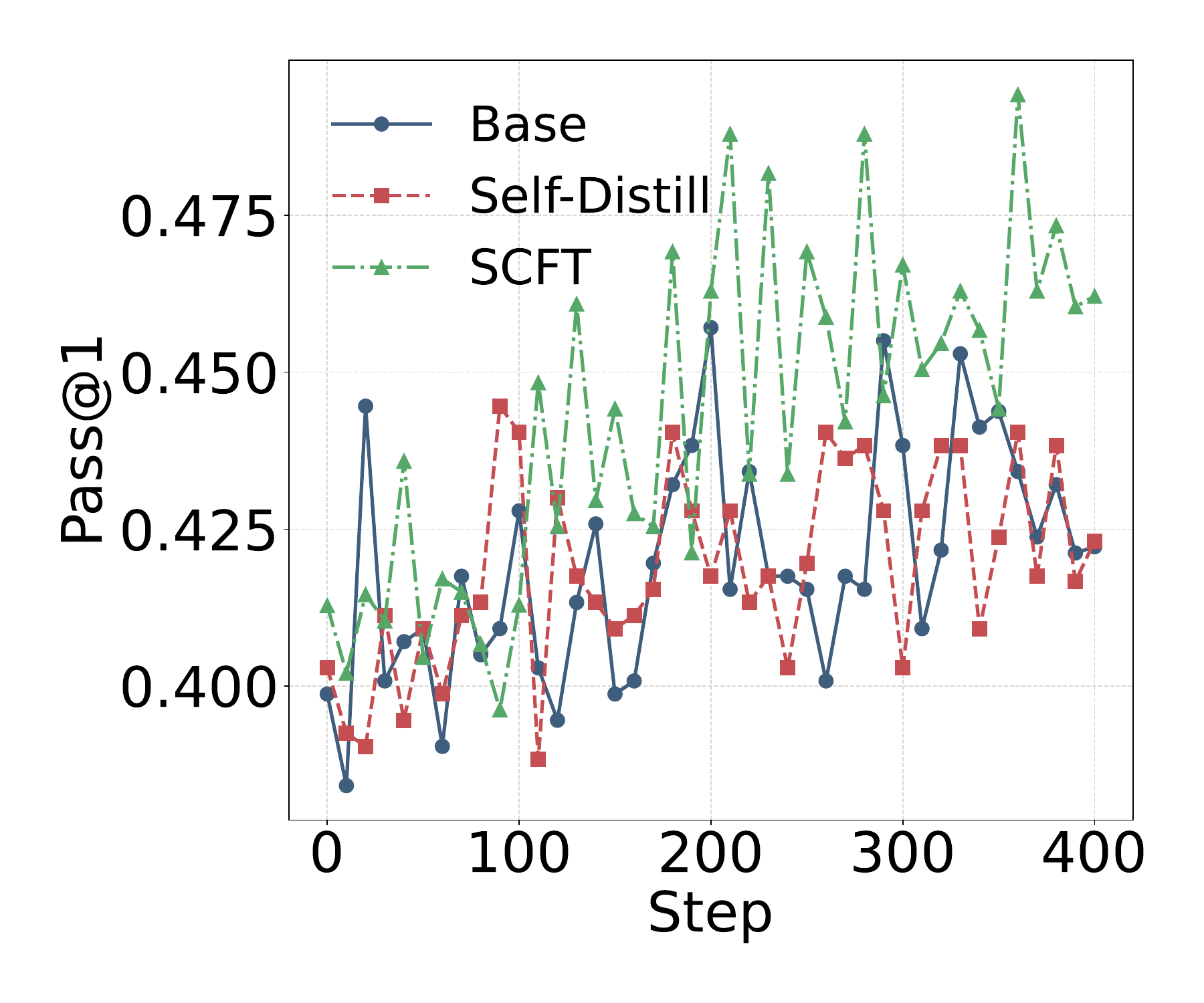}}
    \subfigure[Ablation on Reward Signals.] { \label{subfig:rl2} 
    \includegraphics[width=0.48\linewidth]{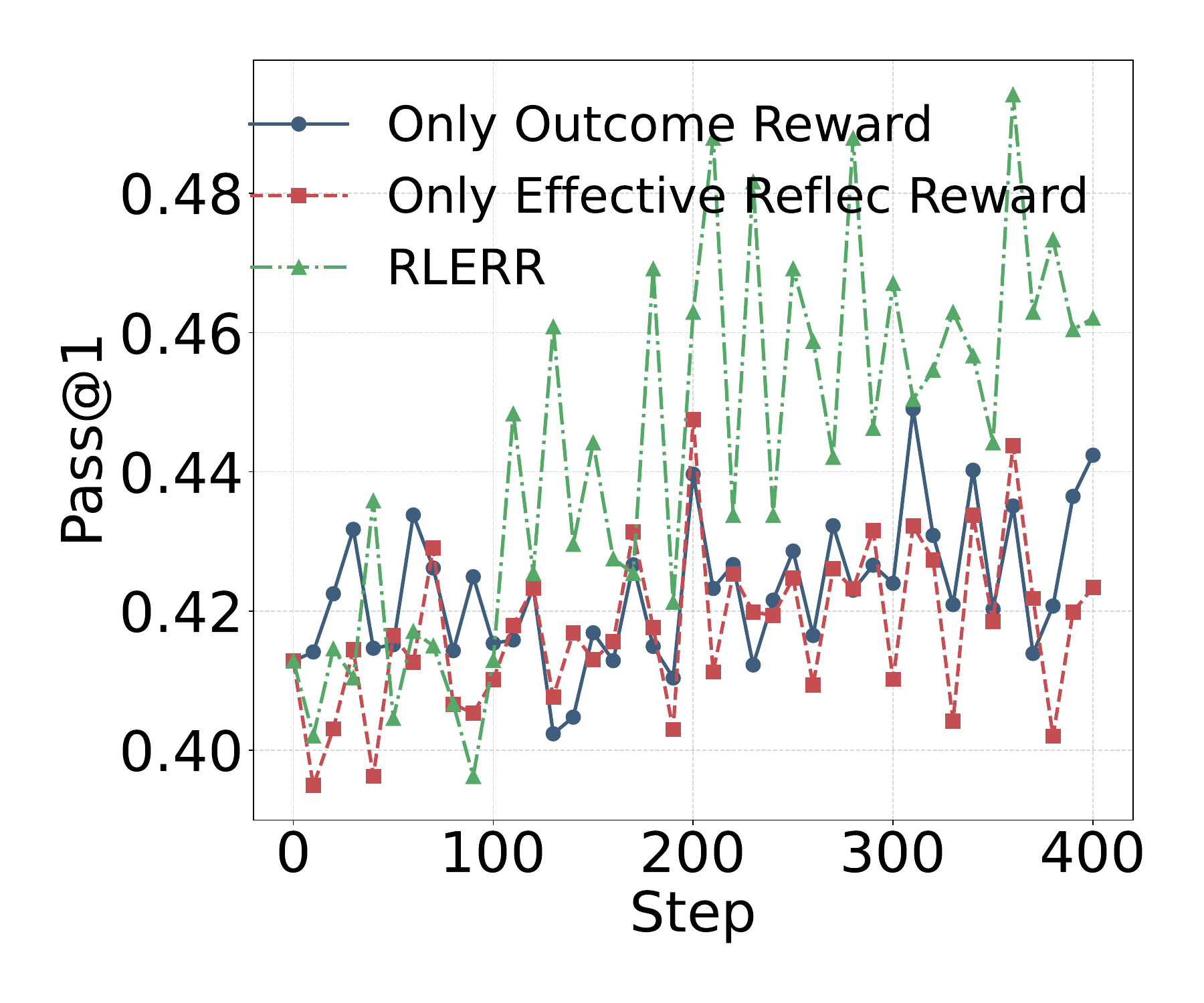}}
        \caption{Analysis of DeepScaleR-1.5B-Preview RL Dynamics on AIME2024. (a) The SCFT-initialized model (green) achieves a higher performance ceiling. (b) RLERR (combining outcome and effective reflection rewards) significantly outperforms using outcome or reflection rewards in isolation.}
    \label{fig:rl}
\end{figure}
\subsection{SCFT Facilitates More Effective RL}
In this subsection, we investigate the training dynamics of Reinforcement Learning with Effective Reflection Rewards (RLERR) under different settings using DeepScaleR-1.5B-Preview.

\textbf{Impact of Initialization.}
As shown in Figure~\ref{subfig:rl1}, the SCFT-initialized model (green) consistently achieves a higher performance ceiling compared to the \textit{Base} and \textit{Self-Distill} baselines, frequently surpassing 47\% Pass@1. This demonstrates that SCFT provides a superior structural prior for self-correction, enabling the model to avoid local optima and explore the solution space more effectively than models that merely mimic correct answers.

\textbf{Ablation on Reward Signals.}
Figure~\ref{subfig:rl2} illustrates the necessity of our hierarchical reward system. While models using only outcome or reflection rewards stagnate around 42-44\% accuracy, the combined RLERR approach (green) achieves a significant boost, peaking near 49\%. This confirms that the synergy between process-oriented reflection rewards and result-oriented outcome rewards is crucial for unlocking the full reasoning potential of LRMs.


\section{Conclusion}



We introduce SCFT and RLERR, a two-stage method designed to transform superficial self-reflections in Large Reasoning Models into effective error-correction mechanisms. By leveraging self-generated critiques via rejection sampling and optimizing with hierarchical reflection rewards, our approach significantly improves both reasoning accuracy and the quality of self-correction. Experiments demonstrate that our method outperforms state-of-the-art baselines, enabling models to autonomously identify and rectify errors without relying on expensive external annotations. 
\section*{Limitations}
Our experiments demonstrate that smaller models (e.g., 1.5B parameters) exhibit limited gains from SCFT. This suggests that models with fewer parameters may lack the capacity to effectively leverage self-critique signals to enhance their reasoning capabilities. Moreover, the rejection sampling process depends on access to ground-truth answers, limiting its applicability in domains where labeled data is scarce. Finally, while we mitigate superficial reflection, there is a risk that the model becomes overly cautious, occasionally attempting to correct already valid reasoning steps.

\section*{Ethics Statement}
This work utilizes publicly available datasets (e.g., AIME, MATH-500) that contain no personally identifiable information or offensive content. Our training objective is grounded in objective mathematical correctness, minimizing the risk of hallucinating harmful content. We acknowledge the environmental impact of computational training costs and commit to releasing our code and models to facilitate reproducibility and reduce redundant efforts.


\bibliography{custom}

\appendix
\section{Appendix}






\subsection{Reflection Evaluation Prompt}
\label{app:evaluation_prompt}
We use the following prompt to evaluate the quality of the model's self-reflection within a complete reasoning trajectory. The evaluator is tasked with identifying reflection moments and scoring their effectiveness based on the defined principles.
\begin{tcolorbox}[
    title=System Prompt for Trajectory-Level Reflection Scoring,
    colback=gray!5!white,
    colframe=gray!75!black,
    breakable, 
    enhanced
]
\textbf{System Prompt:}
You are an expert Mathematics and Logic Evaluator. 
Your task is to assess the quality of the **Self-Reflection/Critique** mechanisms within a model's reasoning trajectory.

\textbf{Input Data:}
\begin{itemize}
    \item \textbf{Question:} [Insert Question]
    \item \textbf{Model Trajectory:} [Insert the full Chain-of-Thought generated by the model]
\end{itemize}
\textbf{Evaluation Process:}
\begin{enumerate}
    \item \textbf{Locate Reflection:} Scan the trajectory for moments where the model pauses to verify, critique, or correct itself (e.g., keywords like ``Wait'', ``Let me double check'', ``Hold on'', or specific tags).
    \item \textbf{Verify Logic:} Independently verify the math/logic at the point of reflection. Determine if the model's previous step was actually correct or incorrect.
    \item \textbf{Apply Principles:} Evaluate the reflection using the 5 Principles ($P_1$-$P_5$).
\end{enumerate}
\textbf{Scoring Rubric (0-10 Scale):}
\begin{itemize}
    \item \textbf{Score 0-2 (Harmful/Hallucinated - $P_1$ Failed):} 
    The model performs a ``Fake Reflection''.
    \begin{itemize}
        \item It claims a correct step is wrong (Hallucination).
        \item It confirms a wrong step is correct (Sycophancy).
        \item \textit{Penalty:} If $P_1$ is violated, the score must be $\le 2$, regardless of length.
    \end{itemize}
    \item \textbf{Score 3-5 (Superficial/Ineffective):} 
    The model attempts to reflect but fails to add value.
    \begin{itemize}
        \item It says ``Let me check'' and simply repeats the same wrong logic.
        \item It identifies an error but fails to fix it ($P_2$ failed).
        \item The reflection is vague (e.g., ``Something is wrong'') without specificity ($P_3$ failed).
    \end{itemize}
    \item \textbf{Score 6-8 (Effective Correction):} 
    The reflection successfully identifies and fixes an error.
    \begin{itemize}
        \item The model correctly spots a calculation or logic error ($P_1, P_3$).
        \item It proposes a valid correction that steers the trajectory back to the right path ($P_2$).
    \end{itemize}
    \item \textbf{Score 9-10 (Profound Insight):} 
    The reflection demonstrates high-level reasoning capabilities.
    \begin{itemize}
        \item It uses a different method to verify the answer (Rigorous Verification - $P_4$).
        \item It breaks a logical loop or provides a shortcut that significantly simplifies the problem (High InfoGain - $P_5$).
    \end{itemize}
\end{itemize}
\textbf{Output Format:}
Return a JSON object with the analysis and final score.
\begin{ttfamily}
\\ \{ \\
\hspace*{1em} "reflection\_moments": ["Wait, I missed the coefficient of x\^{}2 in the expansion..."], \\
\hspace*{1em} "analysis": "The model correctly identified a calculation error.It verified this by re-calculating and provided the correct value, leading to the right answer.", 
\\
\hspace*{1em} "score": 8 \\
\}
\end{ttfamily}
\end{tcolorbox}


\begin{table*}
\centering
\captionsetup{justification=centering}
\resizebox{\linewidth}{!}{
\begin{tabular}{p{\textwidth}}
\toprule

\begin{minipage}{\textwidth}
\underline{\textbf{\small{Instruction:}}}
\tiny{
\textbf{Please critique whether the following solution to the question is correct.}

\textbf{Question:} From an external point \(A\), a tangent \(AB\) and a secant \(ACD\) are drawn to a circle. Find the area of triangle \(CBD\), given that the ratio \(AC : AB = 2 : 3\) and the area of triangle \(ABC\) is 20.

\textbf{Solution:} Okay, so I have this geometry problem where an external point A is drawing a tangent AB and a secant ACD to a circle. The ratio of AC to AB is 2:3, and the area of triangle ABC is 20. I need to find the area of triangle CBD. Hmm, let me think about how to approach this.

First, let me try to visualize the problem. There's a circle...

Given the ratio AC:AB = 2:3. That tells me that AB is longer than AC. Specifically, if AC is 2x, then AB is 3x for some value x. I also know that the area of triangle ABC is 20. I need to find the area of triangle CBD.

...

\[
(3x)^2 = 2x \times AD
\]
\[
9x^2 = 2x \times AD
\]
\[
AD = \frac{9x^2}{2x} = \frac{9x}{2}
\]

Since \(AD = AC + CD\), we find \(CD\):

\[
CD = AD - AC = \frac{9x}{2} - 2x = \frac{9x}{2} - \frac{4x}{2} = \frac{5x}{2}
\]

Next, we use the fact that the areas of triangles sharing the same height are proportional to their bases. Triangles \(ABC\) and \(CBD\) share the same height from vertex \(B\) to the bases \(AC\) and \(CD\) respectively. Therefore, the ratio of their areas is the ratio of their bases \(AC\) and \(CD\):

\[
\text{Ratio of areas} = \frac{AC}{CD} = \frac{2x}{\frac{5x}{2}} = \frac{4}{5}
\]

Given the area of triangle \(ABC\) is 20, the area of triangle \(CBD\) is:

\[
\text{Area of } \triangle CBD = \frac{4}{5} \times 20 = 16  
\]

Thus, the area of triangle \(CBD\) is \(\boxed{16}\).
}

\tiny{- - - - - - - - - - - - - - - - - - - - - - - - - - - - - - - - - - - - - - - - - - - - - - - - - - - - - - - - - - - - - - - - - - - - - - - - - - - - - - - - - - - - - - - - - - - - - - - - - - - - - - - - - - - - - - - - - - - - - - - - - - - - - - - - - - - - - - -}
\\
\underline{\textbf{\small{Response:}}}

<think>
Okay, so I need to evaluate the solution provided for the given math problem.  Let me start by understanding the problem again.

The proposed solution uses the Power of a Point theorem, the ratio of lengths, and then the ratio of areas to conclude that the area of triangle CBD is 16. The user is asking if the solution is correct, if not, to explain why and provide a refined solution.

...

Now, considering triangles ABC and CBD. Both triangles share the same vertex B and their bases AC and CD lie on the same line AD. The key here is that they share the same height from point B to the line AD. Therefore, the ratio of their areas is equal to the ratio of their bases:

\[
\text{Ratio of areas} = \frac{\text{Area of ABC}}{\text{Area of CBD}} = \frac{AC}{CD}
\]

Substituting the values of AC and CD:

\[
\frac{\text{Area of ABC}}{\text{Area of CBD}} = \frac{2x}{\frac{5x}{2}} = \frac{4}{5}
\]

Given that the area of triangle ABC is 20:

\[
\frac{20}{\text{Area of CBD}} = \frac{4}{5}
\]
\[
\text{Area of CBD} = 20 \times \frac{5}{4} = 25
\]

Wait, hold on. My previous conclusion was 16, but upon recalculating using the correct ratio, it's 25. There must be a mistake in the initial reasoning.
...

I realize now that in the initial solution, the ratio was incorrectly interpreted as 4/5, leading to an answer of 16, but in reality, the ratio of AC to CD is 4:5, so CBD's area should be larger than ABC's.

Therefore, the correct area of triangle CBD is 25.

**Final Answer**
The area of triangle \( CBD \) is \boxed{25}.

</think>

Given an external point \(A\) from which a tangent \(AB\) and a secant \(ACD\) are drawn to a circle, we know the ratio \(AC : AB = 2 : 3\) and the area of triangle \(ABC\) is 20. We need to find the area of triangle \(CBD\).

First, we use the Power ...
Next, we use the fact that triangles \(ABC\) and \(CBD\) share the same height from vertex \(B\) to the bases \(AC\) and \(CD\) respectively. Therefore, the ratio of their areas is equal to the ratio of their bases:

\[
\text{Ratio of areas} = \frac{\text{Area of ABC}}{\text{Area of CBD}} = \frac{AC}{CD}
\]

Substituting the values of \(AC\) and \(CD\):

\[
\frac{\text{Area of ABC}}{\text{Area of CBD}} = \frac{2x}{\frac{5x}{2}} = \frac{4}{5}
\]

Given that the area of triangle \(ABC\) is 20:

\[
\frac{20}{\text{Area of CBD}} = \frac{4}{5}
\]
\[
\text{Area of CBD} = 20 \times \frac{5}{4} = 25
\]

Thus, the area of triangle \(CBD\) is \boxed{25}.

\end{minipage} \\
\bottomrule
\end{tabular}
}
\caption{A Training Example.}
\label{tab:training_example}
\end{table*}

\subsection{Example of Training Data}\label{app:example}
As shown in Table \ref{tab:training_example}, we present an example of the training data used in self-critique fine-tuning.

\subsection{Additional Performance Analysis}
\label{app:additional_performance}
\begin{figure*}[t] \centering
    \includegraphics[width=1\textwidth]{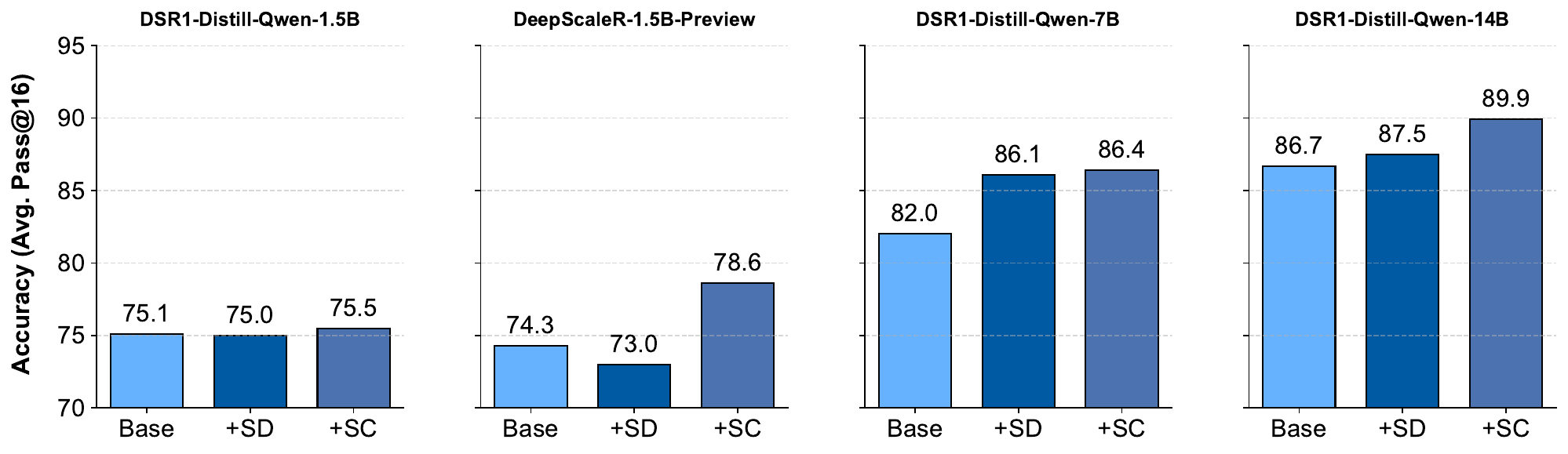}
    \caption{The \texttt{Pass@16} results of different models. ``Base'' represents the model that has not been fine-tuned, and ``SD'' and ``SC'' respectively represent the models fine-tuned through Self-Distill and Self-Critique Fine-tuning.} \label{fig:pass16}
\end{figure*}
In this section, we provide further analysis of the model performance using the \texttt{Pass@16} metric.

As shown in Figure \ref{fig:pass16}, we show \texttt{Pass@16} results for different models. From the results, we can observe that the models fine-tuned with Self-Critique Fine-Tuning (SCFT) consistently outperform their self-distillation counterparts across all datasets. Across models of different scales, SCFT brings an average improvement of 3.1\%. These improvements further demonstrate that SCFT is an effective technique for enhancing model performance.

\subsection{Case Studies}
Finally, we show cases in Table \ref{tab:case} to demonstrate that SCFT+RLERR can teach models reflective behavior in reasoning.
Overall, the response of the model after training can effectively self-reflect within shorter responses and generate correct answers. From the case, we can see that DeepSeek-R1-Distill-Qwen-7B introduces an error ($b=5$) into the originally correct answer ($b=211$). It considers the wrong answer to be correct and uses 32,225 tokens for reasoning. On the contrary, our model can do correct reflection with fewer reasoning tokens and confirm the correct answer. SCFT+RLERR shows its effectiveness in teaching models to identify and correct their own mistakes efficiently.

\begin{table*}[h]
\centering
\captionsetup{justification=centering}
\resizebox{\linewidth}{!}{
\begin{tabular}{p{\textwidth}}
\toprule

\begin{minipage}{\textwidth}
\footnotesize{\textbf{Question (gt = 211):} Let \(b\ge 2\) be an integer. Call a positive integer \(n\) \(b\text-\textit{eautiful}\) if it has exactly two digits when expressed in base \(b\) and these two digits sum to \(\sqrt n\). For example, \(81\) is \(13\text-\textit{eautiful}\) because \(81 = \underline{6} \ \underline{3}_{13} \) and \(6 + 3 = \sqrt{81}\). Find the least integer \(b\ge 2\) for which there are more than ten \(b\text-\textit{eautiful}\) integers.}

\footnotesize{- - - - - - - - - - - - - - - - - - - - - - - - - - - - - - - - - - - - - - - - - - - - - - - - - - - - - - - - - - - - - - - - - - - - - - - - - - - - - - - - - - - - - - - - - -}
\textbf{\small{DeepSeek-R1-Distill-Qwen-7B:}} \footnotesize{Okay, so I need to find the least integer \( b \geq 2 \), ... Hence, \( b -1 \) must have at least four distinct prime factors. The smallest such \( b -1 \) is \( 2 \times 3 \times 5 \times 7 = 210 \), so \( b = 211 \). \correct

But wait, \textcolor{blue}{let's confirm for \( b = 211 \), \( b -1 = 210 \).} The number of solutions \( s \) is \( 2^4 =16 \). So, 16 numbers. But wait, in our earlier case, for \( b =13 \), which had \( b -1 =12 \), which has two prime factors, 2 and 3, we had two solutions...

\textcolor{red}{Wait, this is conflicting. So, maybe my formula is incorrect}... So, let's compute \( \dbinom{b^2 +b -1}{b -1} \)for increasing \( b \) until it exceeds1000. Compute for \( b=2 \):
... \( b=5 \):... Wait, hold on, let me compute: \( b=5 \): \( b^2 +b -1=25 +5 -1=29 \). Compute \( \dbinom{29}{4} = \frac{29*28*27*26}{4*3*2*1} = \frac{29*28*27*26}{24} \)...  So, \( \dbinom{29}{4}=23751 \), which is greater than1000.

Therefore, for \( b=5 \), the number exceeds1000. But wait, is \( b=5 \) the minimal? let me check \( b=4 \): 969 <1000. So, the minimal \( b \) is 5\wrong... \textcolor{red}{But let me verify for \( b=4 \):969<1000, so indeed, \( b=5 \) is the minimal.\wrong} \textbf{(32225 tokens)}}

- - - - - - - - - - - - - - - - - - - - - - - - - - - - - - - - - - - - - - - - - - - - - - - - - - - - - - - - - - - - - - - - - - - - - - - - - - - - - - - - - - - - - - - - - -

\textbf{\small{Our Model:}} \footnotesize{Okay, so I need to solve this math problem  ...  Let me break this down step by step. First, ... But the problem says "the least integer \( b \geq 2 \) for which there are more than ten \( b \)-eautiful integers." \textcolor{blue}{So, 211 might be the answer \correct , but that seems too large, and the problem is likely expecting a smaller answer.}

Wait, perhaps my entire approach is wrong. \textcolor{blue}{Let me think again.
let me verify} for \( b=13 \). Earlier, I found 3 solutions, but perhaps I missed some. s=4,9,12. Is that all? Check s=3: 3*2=6. 12 divides 6? No. s=5:5*4=20. 12 divides 20? ...   So, only 3 solutions. So, with \( \omega=2 \), only 3 solutions, which is less than 7 with \( \omega=3 \). So, it's not a linear relationship. Therefore, the number of solutions doesn't scale exactly with \( \omega \). \textcolor{ForestGreen}{Hence, my conclusion that \( b=211 \) is correct, despite it being large. \correct } \textbf{(14556 tokens)}}

\end{minipage} \\
\bottomrule
\end{tabular}
}
\caption{Case Studies. Our model undergoes effective reflection and provides the correct answer. The \textcolor{blue}{blue} text represents reflection, and the \textcolor{ForestGreen}{green}/\textcolor{red}{red} text respectively represent obtaining the correct/wrong answers.}
\label{tab:case}
\end{table*}

\end{document}